\documentclass[runningheads,a4paper]{llncs}
\usepackage{verbatim}
\usepackage{wrapfig}
\usepackage[table,xcdraw]{xcolor}
\usepackage{multirow}
\usepackage{amssymb}
\usepackage{graphicx}
\usepackage{rotating}
\usepackage{adjustbox}
\usepackage[utf8]{inputenc}
\usepackage{url}
\usepackage{amsmath}
\usepackage{enumitem}

\begin{document}

\mainmatter  % start of an individual contribution

\title{\emph{CardiacNET}: Segmentation of Left Atrium and Proximal Pulmonary Veins from MRI Using Multi-View CNN}

\titlerunning{CardiacNET}

%\author{Aliasghar Mortazi, Jeremy Burt$^{\dagger}$, Ulas Bagci$^{\star}$}
%\thanks{Thanks to NVidia for donating a GPU for deep learning experiments.}%


%\institute{$^{\star}$ Center for Research in Computer Vision (CRCV), University of Central Florida, Orlando, FL.\\$^{\dagger}$ Diagnostic Radiology Department, Florida Hospital, Orlando, FL.}

\author{Aliasghar Mortazi$^{1}$,Rashed Karim$^{2}$, Kawal Rhode$^{2}$, \\Jeremy Burt$^{3}$, Ulas Bagci$^{1}$}
\authorrunning{Aliasghar Mortazi, Rashed Karim, Kawal Rhode, Jeremy Burt, Ulas Bagci}
\institute{$^{1}$ Center for Research in Computer Vision (CRCV), University of Central Florida, Orlando, FL.\\$^{2}$Division of Imaging Sciences and Biomedical Engineering, King's College London, London, UK \\$^{3}$ Diagnostic Radiology Department, Florida Hospital, Orlando, FL.\\ \emph{E-mail: a.mortazi@knights.ucf.edu}}
\toctitle{Lecture Notes in Computer Science}
\tocauthor{Authors' Instructions}
\maketitle

\begin{abstract}
Anatomical and biophysical modeling of left atrium (LA) and proximal pulmonary veins (PPVs) is important for clinical management of several cardiac diseases. Magnetic resonance imaging (MRI) allows qualitative assessment of LA and PPVs through visualization. However, there is a strong need for an advanced image segmentation method to be applied to cardiac MRI for quantitative analysis of LA and PPVs. In this study, we address this unmet clinical need by exploring a new deep learning-based segmentation strategy for quantification of LA and PPVs with high accuracy and heightened efficiency. Our approach is based on a multi-view convolutional neural network (CNN) with an adaptive fusion strategy and a new loss function that allows fast and more accurate convergence of the backpropagation based optimization. After training our network from scratch by using more than 60K 2D MRI images (slices), we have evaluated our segmentation strategy to the STACOM 2013 cardiac segmentation challenge benchmark. Qualitative and quantitative evaluations, obtained from the segmentation challenge, indicate that the proposed method achieved the state-of-the-art sensitivity (90\%), specificity (99\%), precision (94\%), and efficiency levels (10 seconds in GPU, and 7.5 minutes in CPU). %We have also tested our proposed method on an independent cohort that include 20 test MRI, and similar segmentation accuracies were obtained (99\%).
\end{abstract}

\section{Introduction}
Atrial fibrillation (AF) is a cardiac arrhythmia caused by abnormal electrical discharges in the atrium, often beginning with hemodynamic and/or structural changes in the left atrium (LA)~\cite{Kuppahally-circulation}. AF is clinically associated with LA strain, and MRI is shown to be a promising imaging method for assessing the disease state and predicting adverse clinical outcomes.  The LA also has an important role in patients with ventricular dysfunction as a booster pump to augment ventricular volume~\cite{daoudi}. Computed tomography (CT) imaging of the heart is frequently performed  when managing AF and prior to pulmonary vein ablation (isolation) therapy due to its rapid processing time. In recent years, there is an increasing interest in shifting towards cardiac MRI due to its excellent soft tissue contrast properties and lack of radiation exposure.  For pulmonary vein ablation therapy planning in AF,  precise segmentation of the LA and PPVs is essential. However, this task is non-trivial because of multiple anatomical variations of LA and PPV. 

Historically, statistical shape and atlas-based methods have been the state-of-the-art cardiac segmentation approaches due to their ability to handle large shape/appearance variations. One significant challenge for such approaches is their limited efficiency: an average of 50 minutes processing time per volume~\cite{Karim}. Statistical shape models are faster than atlas-based methods, and a high degree uncertainties in the accuracy of such models is inevitable~\cite{Stender}. To alleviate this problem and accomplish the segmentation of LA and PPVs from 3D cardiac MRI with high \emph{accuracy} and \emph{efficiency}, we propose to a new deep CNN. Our proposed method is fully automated, and largely different from previous methods of LA and PPVs segmentation. The summary of these differences and key novelties of the proposed method, named as \emph{CardiacNET}, are listed as follows:

%\noindent\textbf{Summary of our contributions.} We present a fully automated LA and PPVs segmentation strategy, based on a newly designed multi-view deep CNN architecture. Our work has a number of novelties and substantial differences from the currently used cardiac segmentation methods as follows:
\let\labelitemi\labelitemii
\begin{itemize}
\item  Training CNN from scratch for 3D cardiac MRI is not feasible with insufficient 3D training data (with ground truth) and limited computer memory. Instead, we parsed 3D data into 2D components (axial (A), sagittal (S), and coronal (C)), and utilized a separate deep learning architecture for each component. The proposed \emph{CardiacNET} was trained using more than 60K 2D slices of cardiac MR images without relying on a pre-training network of non-medical data.
\item We have combined three CNN networks through an adaptive fusion mechanism where complementary information of each CNN was utilized to improve segmentation results. The proposed adaptive fusion mechanism is based on a new strategy; called \textit{robust region}, which measures (roughly) the reliability of segmentation results without the need for ground truth.
\item We devised a new loss function in the proposed network, based on a modified z-loss, to provide fast convergence of network parameters. This not only improved segmentation results due to fast and reliable allocation of network parameters, but it also provided a significant acceleration of the segmentation process. The overall segmentation process for a given 3D cardiac MRI takes at most 10 seconds in GPU, and 7.5 minutes in CPU on a normal workstation.
\end{itemize}

%Experimental results indicated that our proposed approach has achieved the-state-of-the-art segmentation results in accuracy and significantly better efficiency reported both on CPU and GPU, showing the premise of future research possibilities for auto-analysis of other substructure of heart.    

\begin{figure}[t]
\vspace{-0.5 cm}
\centering
\includegraphics[height=4cm,width=1\textwidth]{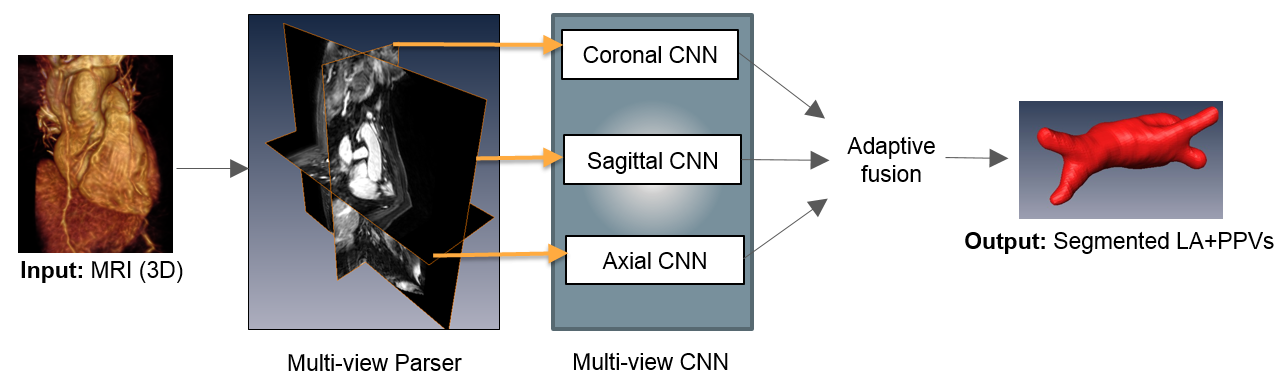}
\caption{High-level overview of the proposed multi-view CNN architecture.}
\vspace{-0.5 cm}
\end{figure}

\section{Proposed Multi-View Convolutional Neural Network (CNN) Architecture}
The proposed pipeline for deep learning based segmentation of the LA and PPVs is summarized in Fig.1. We used the same CNN architecture for each view of the 3D cardiac MRI after parsing them into axial, sagittal, and coronal views. The rationale behind this decision is based on the limitation of computer memory and insufficient 3D data for training on 3D cardiac MRI from scratch. Instead, we reduced the computational burden of the CNN training by constraining the problem into a 2D domain. The resulting pixel-wise segmentations from each CNN are combined through an adaptive fusion strategy. The fusion operation was designed to maximize the information content from different views. The details of the pipeline are given in the following subsections.\\

\noindent\textbf{Encoder-Decoder CNN:} We constructed an encoder-decoder CNN architecture, similar to that of Noh et al.~\cite{Noh}. The network includes 23 layers (11 in encoder, 12 in decoder units). Two max-pooling layers in encoder units reduce the image dimensions by half, and a total of 19 convolutional (9 in encoder, 10 in decoder), 18 batch normalization, and 18 ReLU (rectified linear unit) layers are used. Specific to the decoder unit, two upsampling layers are used to convert the images back into original sizes. Also, the kernel size of all filters are considered as $3\times3$. The final layer of the network includes a softmax function (logistic) for generating a probability score for each pixel. Details of these layers, and associated filter size and numbers are given in Fig.2.\\

\noindent\textbf{Loss Function:} We used a new loss function that can estimate the parameters of the proposed network at a much faster rate. We trained end-to-end mapping with a loss function $L(\mathbf{o},c)=$softplus$(a(b-z_c))/a$, called z-loss~\cite{z-loss}, where $\mathbf{o}$ denotes output of the network, $c$ denotes the ground truth label, and $z_c$ indicate z-normalized label, obtained as $z_c=(o_c-\mu)/\sigma$ where mean ($\mu$) and standard deviation $\sigma$ are obtained from $\mathbf{o}$. z-loss is simply obtained with the reparametrization of \emph{soft-plus} (SP) function (i.e., $SP(x)=ln(1+e^x)$) through two hyperparameters: $a$ and $b$. Herein, we kept these hyperparameters fixed, and trained the network with a reduced z-loss function. The rationale behind this choice is the following: the z-loss function provides an efficient training performance as it belongs to spherical loss family, and it is invariant to scale and shift changes in the output, avoiding output parameters to deviate from extreme values.\\

%In addition to widely used mean square error (MSE) and cross entropy based loss functions, 
%learning rate of 0.00001  has been used to minimized the loss function.\\ 

\noindent\textbf{Training \emph{CardiacNET} from scratch:} 3D cardiac MRI images along with its corresponding expert annotated ground truths were used to train the CNN after the images are parsed into three views (A, S, C). Data augmentation has been conducted on the training dataset with translation and rotation operation as indicated in Table 1. Obtained 3D images were parsed into A, S, and C views, and more than 60K 2D images were obtained to feed training of the CNN (approximately 30K for A and C views, around 11K for S view). As a preprocessing step, all images have undergone anisotropic smoothing filtering and histogram matching.\\

\begin{figure}[t]
\centering
\begin{turn}{0}
\centering
\includegraphics[width=1\textwidth]{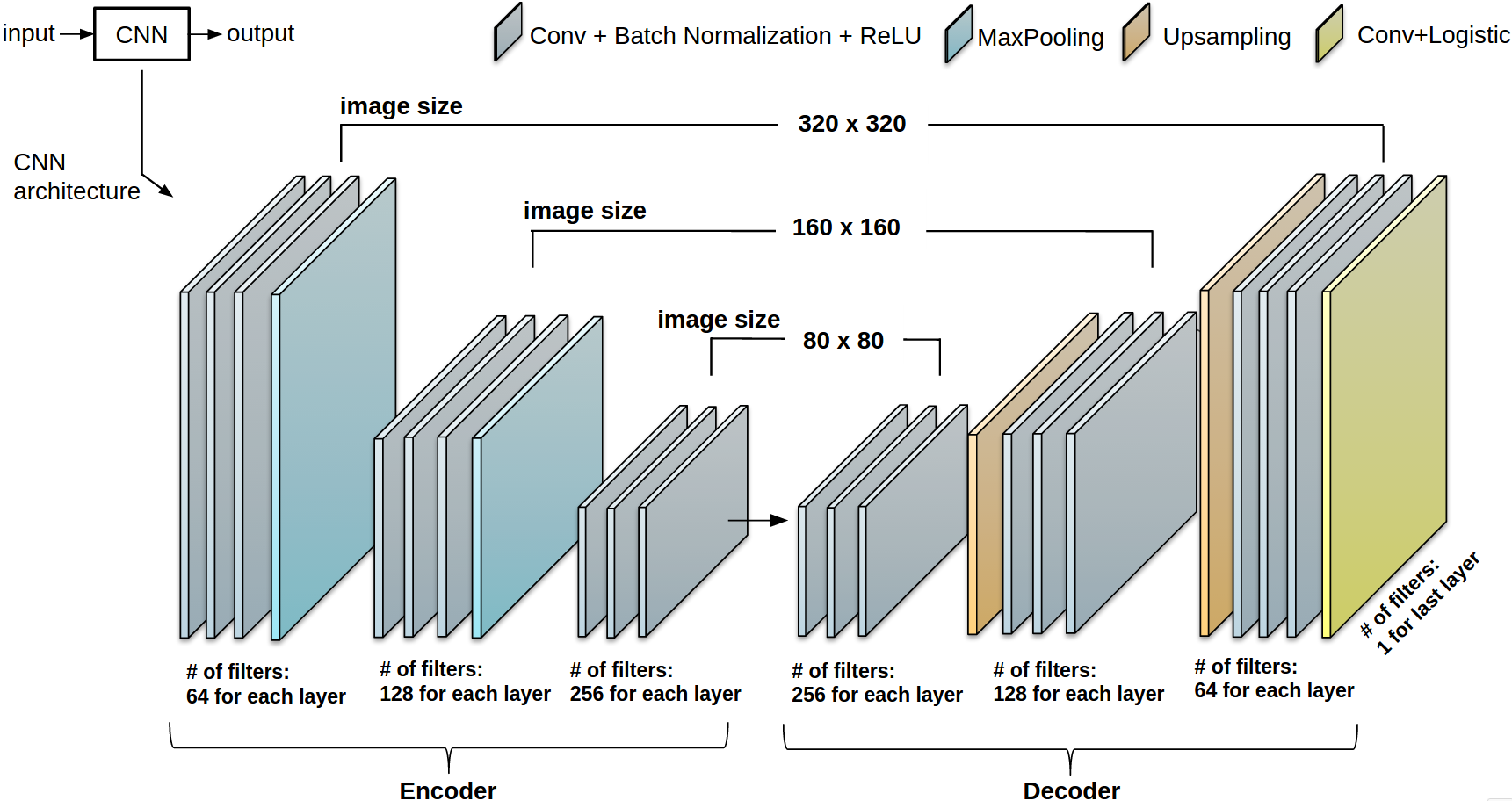}
\end{turn}
\caption{Details of the CNN architecture. Note that image size is not necessarily fixed for each view's CNN.}
\vspace{-0.6 cm}
\end{figure}
\vspace{-0.3 cm}

\begin{wraptable}{r}{0.49\textwidth}
\vspace{-1 cm}
\caption{Data augmentation parameters and number of training images}
\vspace{-0.5 cm}
\begin{adjustbox}{max width=0.5\textwidth}
\label{my-label}

\begin{tabular}{lllllccclll}
 &  &  &  &                       & \multicolumn{1}{l}{}                                & \multicolumn{1}{l}{}                       & \multicolumn{1}{l}{}                     &  &  &  \\
 &  &  &  &                       & \multicolumn{1}{l}{}                                & \multicolumn{1}{l}{}                       & \multicolumn{1}{l}{}                     &  &  &  \\ \cline{6-8}
 &  &  &  & \multicolumn{1}{l|}{} & \multicolumn{3}{c|}{\cellcolor[HTML]{C0C0C0}\textit{\textbf{Data augmentation}}}                                                            &  &  &  \\ \cline{6-8}
 &  &  &  & \multicolumn{1}{l|}{} & \multicolumn{1}{c|}{\textbf{Methods}}               & \multicolumn{2}{c|}{\textbf{Parameters}}                                              &  &  &  \\ \cline{6-8}
 &  &  &  & \multicolumn{1}{l|}{} & \multicolumn{1}{c|}{}                               & \multicolumn{2}{c|}{{{$(x+\textit{trans},y=0),\textit{trans}\, \varepsilon [-20,20]$ }}}                                                               &  &  &  \\
 &  &  &  & \multicolumn{1}{l|}{} & \multicolumn{1}{c|}{\multirow{-2}{*}{Translations}} & \multicolumn{2}{c|}{{$(x=0,y+\textit{trans}),\textit{trans}\, \varepsilon [-20,20]$ }}                                                               &  &  &  \\ \cline{6-8}
 &  &  &  & \multicolumn{1}{l|}{} & \multicolumn{1}{c|}{Rotation}                       & \multicolumn{2}{c|}{{$k\times45, k\, \varepsilon [-2,-1,1,2]$ }}                                                               &  &  &  \\ \cline{6-8}
 &  &  &  & \multicolumn{1}{l|}{} & \multicolumn{3}{c|}{\cellcolor[HTML]{C0C0C0}\textit{\textbf{Training images}}}                                                              &  &  &  \\ \cline{6-8}
 &  &  &  & \multicolumn{1}{l|}{} & \multicolumn{1}{c|}{\textbf{CNN}}                   & \multicolumn{1}{c|}{\textbf{\# of images}} & \multicolumn{1}{c|}{\textbf{Image size}} &  &  &  \\ \cline{6-8}
 &  &  &  & \multicolumn{1}{l|}{} & \multicolumn{1}{c|}{Sagittal}                       & \multicolumn{1}{c|}{10,800}                & \multicolumn{1}{c|}{320 $\times$ 320}           &  &  &  \\
 &  &  &  & \multicolumn{1}{l|}{} & \multicolumn{1}{c|}{Axial}                          & \multicolumn{1}{c|}{28,800}                & \multicolumn{1}{c|}{110 $\times$ 320}           &  &  &  \\
 &  &  &  & \multicolumn{1}{l|}{} & \multicolumn{1}{c|}{Coronal}                        & \multicolumn{1}{c|}{28,800}                & \multicolumn{1}{c|}{110 $\times$ 320}           &  &  &  \\ \cline{6-8}
\end{tabular}
\end{adjustbox}
\vspace{-0.5 cm}
\end{wraptable}

%More recently, machine learning based methods, specifically deep learning methods~\cite{}, have been shown effective in solving many computer vision tasks such as classification, regression, and segmentation. This has been observed in medical image analysis applications too: lung cancer detection and diagnosis~\cite{} and classification of brain diseases from MRI are only a few to name. Despite these advances, machine learning based approaches are inferior in LA and PPVs segmentation, potentially due to sub-optimal design and  accuracies are not superior to 

\noindent\textbf{Multi-View Information Fusion.} Since cardiac MRI is often not reconstructed with isotropic resolution, we expected varying segmentation accuracy in different views. In order to alleviate potential adverse effects caused by non-isotropic spatial resolutions of a particular view, it is desirable to reduce the contribution of that view into final segmentation. We have achieved this with the adaptive fusion strategy as described next. For a given MRI volume \textbf{I}, and its corresponding segmentation $\mathbf{o}$, we proposed a new strategy, called \emph{robust region}, that roughly determined the reliability of the output segmentation $\mathbf{o}$ by assessing its object distribution. To achieve this, we hypothesized that the output should include only one connected object when the segmentation is successful, and if there was more than a single connected object available, these can be considered as false positives. Accordingly, respective performance of segmentation performance in A, S, and C views can be compared and weighted. To this end, we utilized connected component analysis (CCA) to rank output segmentations and reduced the contribution of CNN for a particular view when false positive findings (non-trusted objects/components) were large and true positive findings (trusted object/component) were small. Fig.3 describes the adaptive fusion strategy as $CCA (\mathbf{o})=\{o_{1},\ldots,o_{n} | \cup o_{i} = \mathbf{o}, \text{ and } \cap o_i=\phi\}$. Thus, the contribution of each view's CNN was computed based on a weighting $ w= {max_{i} \{|o_{i}| \}}/  \sum_{i} |o_{i}| $, indicating that higher weights were assigned when the component with largest volume dominated the whole output volume. Note that this block has been used only in the test phase. Complementary to this strategy, we also used simple linear fusion of each views for comparison (See Experimental Results section).
\vspace{-0.5 cm}
\begin{figure}[h!]
\centering
\includegraphics[width=1\textwidth]{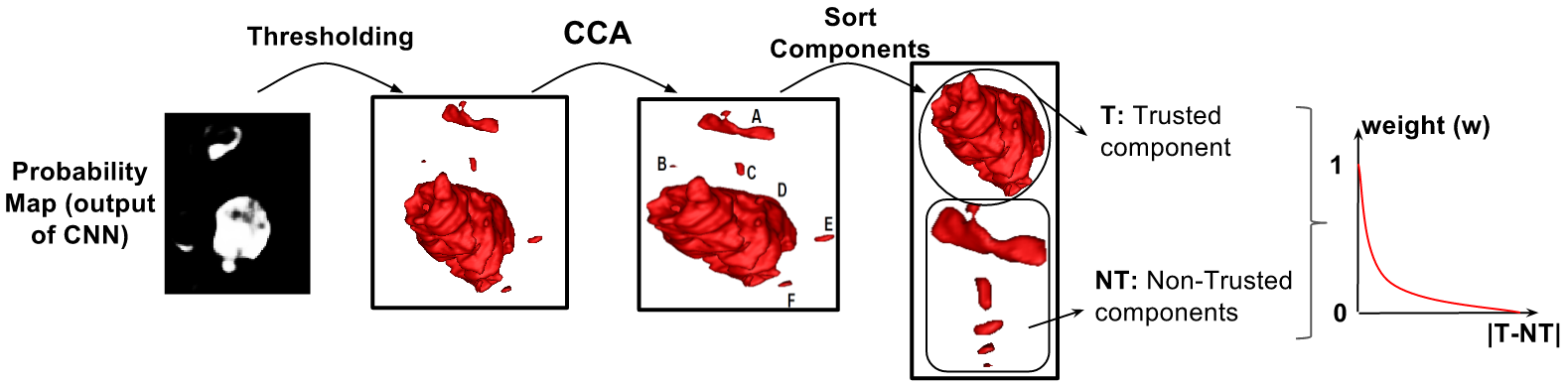}
\caption{Connected components obtained from each view were computed and the residual volume (T-NT) was used to determine the strength for fusion with the other views.}
\end{figure}
\vspace{-0.7 cm}

\section{Experimental Results}
\vspace{-0.3 cm}
\textbf{Data sets:} Thirty cardiac MRI data sets were provided by the STACOM 2013 challenge organizers~\cite{Karim}. Ten training data were provided with ground truth labels, and the remaining twenty were provided as a test set. It is important to note that not the complete PVs are considered in the segmentation challenge, but only the proximal segments of the PVs up to the first branching vessel or after 10 mm from the vein ostium were included in the segmentation. MR images were obtained from a 1.5T Achieva (Philips Healtcare, The Neatherlands) scanner with an ECG-gated 3D balanced steady-state free precession acquisition~\cite{Karim} with TR/TE$=4.4/2.4$ ms, and Flip-angle=$90^o$. Typical acquisition time for the cardiac volume imaging was 10 minutes. In-plane resolution was recorded as 1.25 $\times$ 1.25 mm$^2$, slice thickness was measured as 2.7 mm. Further details on the data acquisition, and image properties can be found in~\cite{Karim}.\\

\noindent\textbf{Evaluations.} For evaluation and comparison with other state-of-the-art method, we have used the same evaluation metrics, provided by the STACOM 2013 challenge: Dice index and surface-to-surface (S2S) metrics. In addition, we calculated Dice index and S2S for the LA and PPVs separately. To provide a comprehensive evaluation and comparisons, sensitivity (true positive rate), specificity (true negative rate), precision (positive prediction value), and Dice index values for the combined LA and PPVs were included too. Table 2 summarizes all these evaluation metrics along with efficiency comparisons where we tested our algorithm both in GPU and CPU. LTSI-VRG, UCL-1C, and UCL-4C are three atlas-based method which their output were published publicly as a part of STACOM 2013 challenge. Also, OBS-2 is the result from human observer which its output was available as a part of STACOM 2013 challenge.  Using leave-one-out cross-validation strategy on training dataset, we achieved high sensitivity (0.92) and Dice value (0.93). Similarly, in almost all evaluation metrics in the test set, the proposed method out-performed the state-of-the-art approaches by large margins. Table 2 indicates the results of varying combinations using \emph{CardiacNET} such as single CNN in particular view (i.e,. $S_CNN$), with simple linear fusion F-CNN, adaptive fusion AF-CNN, and with the new loss function AF-CNN-SP. In AF-CNN, the loss function was cross-entropy. The best method in the challenge data set was reported to have a Dice index of 0.94 for LA and 0.65 for PPVs (combined LA and PPVs was less than 0.9). In our proposed method, the Dice index for combined LA and  PPVs was well above 0.90. For efficiency comparison, our approach only takes at most 10 seconds on a Nvidia TitanX GPU and 7.5 minutes in a CPU with Octa-core processor (2.4 GHz) configuration. The method in~\cite{UCL} required 30-45 minutes of processing times (with Quad-core processor (2.13 GHz)). For qualitative evaluation, we have used surface rendering of output segmentations compared to ground truth in Fig.4. Sample axial, sagittal, and coronal MRI slices are given in the same figure with ground truth annotations overlaid with the segmented LA and PPVs.

\begin{comment}
\begin{figure}[h!]
\centering
\includegraphics[height=2.5cm,width=1\textwidth]{P1_2ds.png}
\includegraphics[height=2.5cm,width=1\textwidth]{P4_2ds.png}
\includegraphics[height=2.5cm,width=1\textwidth]{P10_2ds.png}
\caption{Proposed method delineation results (green) with respect to the ground-truth (red). Boundaries are given for three different subjects at three different slice. Right corner of each image shows zoomed LAs. }
\end{figure}
\end{comment}

\begin{figure}[h!]
\vspace{-0.5 cm}
\centering
\includegraphics[height=2.5 cm,width=0.7\textwidth]{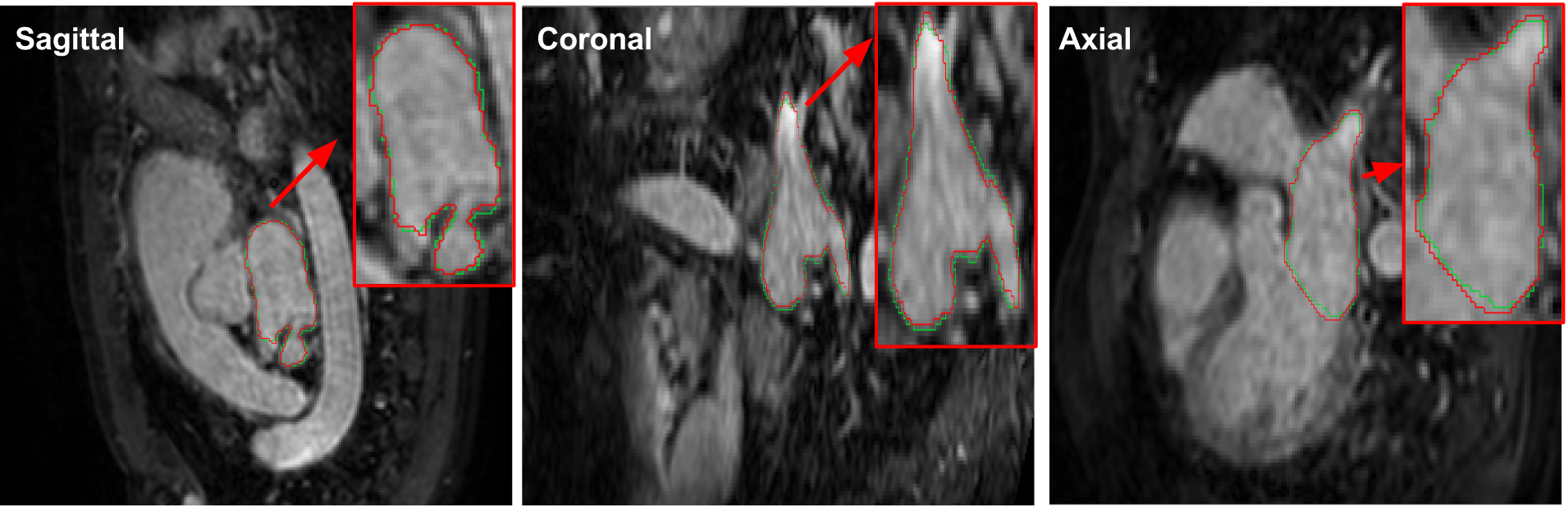}
\\
$\:$
\includegraphics[height=2 cm,width=0.7\textwidth]{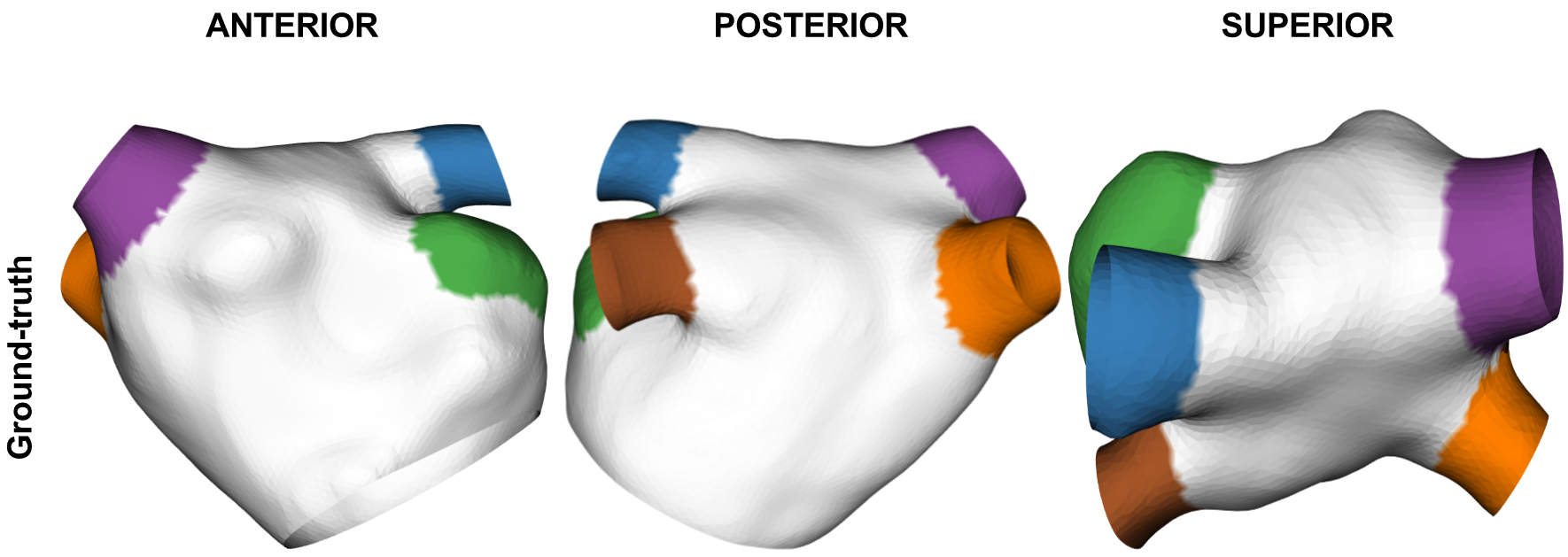}
\includegraphics[height=2 cm,width=0.7\textwidth]{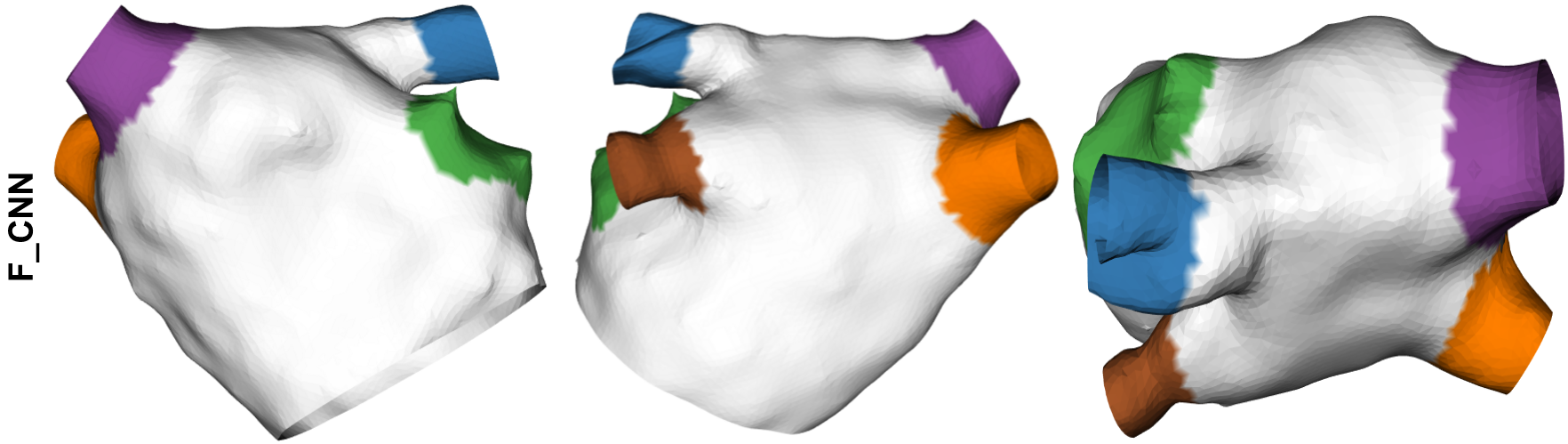}
\includegraphics[height=2 cm,width=0.7\textwidth]{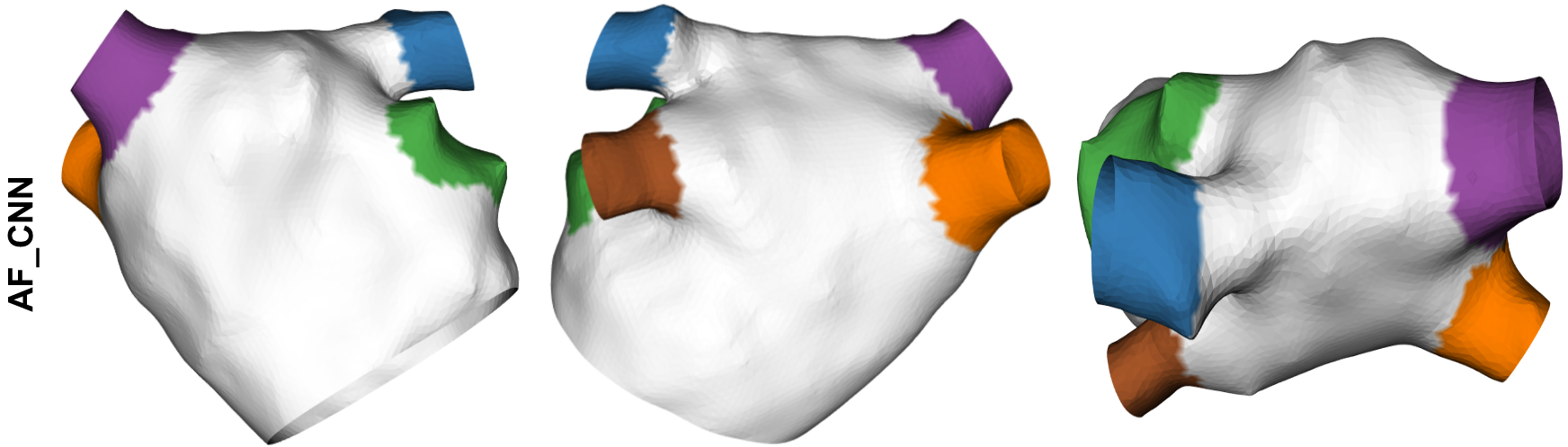}
\includegraphics[height=2 cm,width=0.7\textwidth]{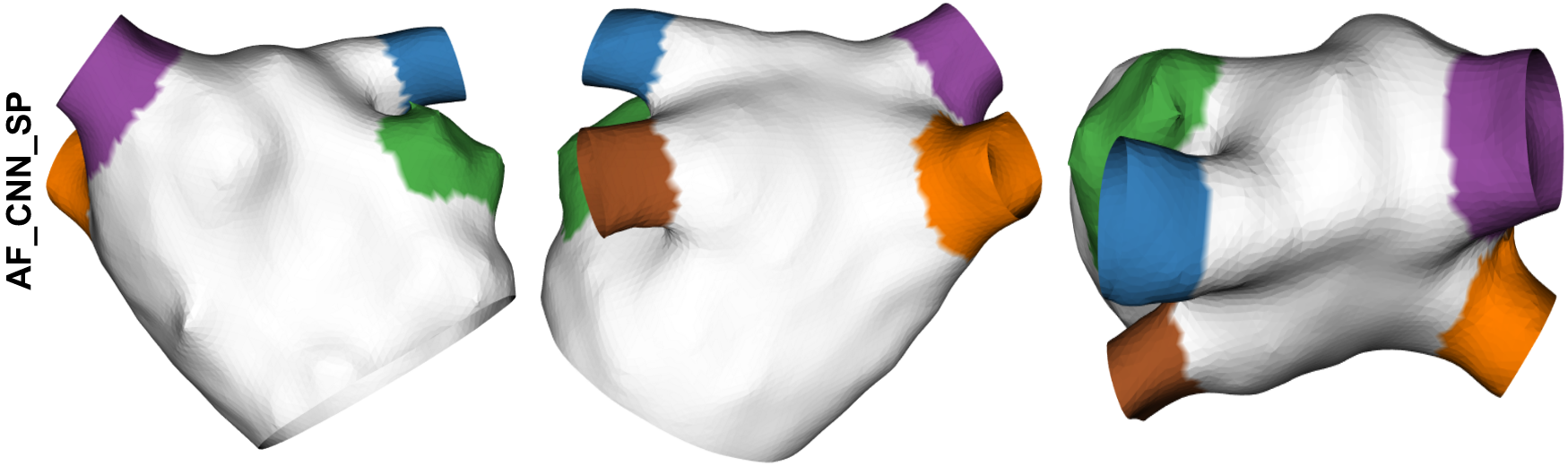}
\caption{First row shows sample MRI slices from S, C, and A views (red contour is ground-truth and green one is output of proposed method). Second-to-fifth rows: 3D surface visualization for the ground-truth and the output generated by the proposed method w.r.t simple fusion (F), adaptive fusion (AF), and the new loss function (SP).}
\vspace{-0.6 cm}
\end{figure}

\begin{table}
\vspace{-0.4 cm}
\caption{\label{tab:table-name} The evaluation metrics for state-of-the-art and proposed methods. $^{**}$: the running time on CPU $^*$: the running time on NVIDIA TitanX GPU}
\centering
 \begin{adjustbox}{max height=14 cm,max width=1\textwidth}
\begin{tabular}{|c|cccccccccc|}
    \hline 
  \textbf{Methods}    & \textbf{LTSI\_VRG} & \textbf{UCL\_1C} & \textbf{UCL\_4C} 
  & \textbf{OBS\_2} & \textbf{A\_CNN} & \textbf{C\_CNN} &  \textbf{S\_CNN}  & \textbf{F-CNN}  &  \textbf{AF-CNN} &  \textbf{AF-CNN-SP} \\ \hline 
    
     \textbf{Dice(LA)} & 0.910& 0.938 & 0.859 & 0.908 & 0.903 & 0.804 & 0.787 & 0.873 & 0.928 & \textbf{0.951}\\

     \textbf{Dice(PPVs)} & 0.653&	0.609	&0.646	&0.751	&0.561	&0.478&	0.398&	0.506&	0.616 & \textbf{0.685}\\
    
     \textbf{S2S(LA) in mm} & 1.640 & 1.086 & 2.136 & 1.538 & 1.592 & 2.679 & 2.853 & 1.771 & 1.359  & \textbf{1.045}\\

      \textbf{S2S(PPVs) in mm} & 1.994 & 1.623 & 2.375 & 1.594 & 1.928 & 2.878 & 3.581 & 2.121 & 1.718 & \textbf{1.427}\\ 
\hline
     \textbf{Sensitivity} & \textbf{0.926} & 0.828 & 0.832 & 0.894 & 0.806 & 0.658 & 0.663 & 0.743 & 0.883 & 0.895\\
     \textbf{Specificity} & 0.998 & 0.999 & 0.999 & 0.997 & 0.996 & 0.994 & 0.997 & 0.997 & 0.999 & \textbf{0.999}\\
     \textbf{Precision} & 0.815 & \textbf{0.957} & 0.814 & 0.936 & 0.905 & 0.774 & 0.880 & 0.953 & 0.936 & 0.938\\
     \textbf{Dice (all) }& 0.862 & 0.886 & 0.819 & 0.911 & 0.845 & 0.695 & 0.734 & 0.820 & 0.887 & \textbf{0.905}\\
     \hline
    \textbf{Running}& 3100$^{**}$ & 1200$^{**}$ & 1200$^{**}$ & - & $170^{**}$ & $170^{**}$ & $155^{**}$ & \textbf{450}$^{**}$ & \textbf{450}$^{**}$ & \textbf{450}$^{**}$\\
  \textbf{Time (sec)}  & - & - & - & - & $3.5^{*}$ & $3.5^{*}$ & $3^{*}$ & \textbf{10$^{*}$}& \textbf{10$^{*}$} & \textbf{10$^{*}$}\\
    \hline
\end{tabular}
\end{adjustbox}
\vspace{-1 cm}
\end{table}

\begin{figure}[h!]
\vspace{-0.4 cm}
\centering
\begin{turn}{0}
\centering
\includegraphics[width=1\textwidth]{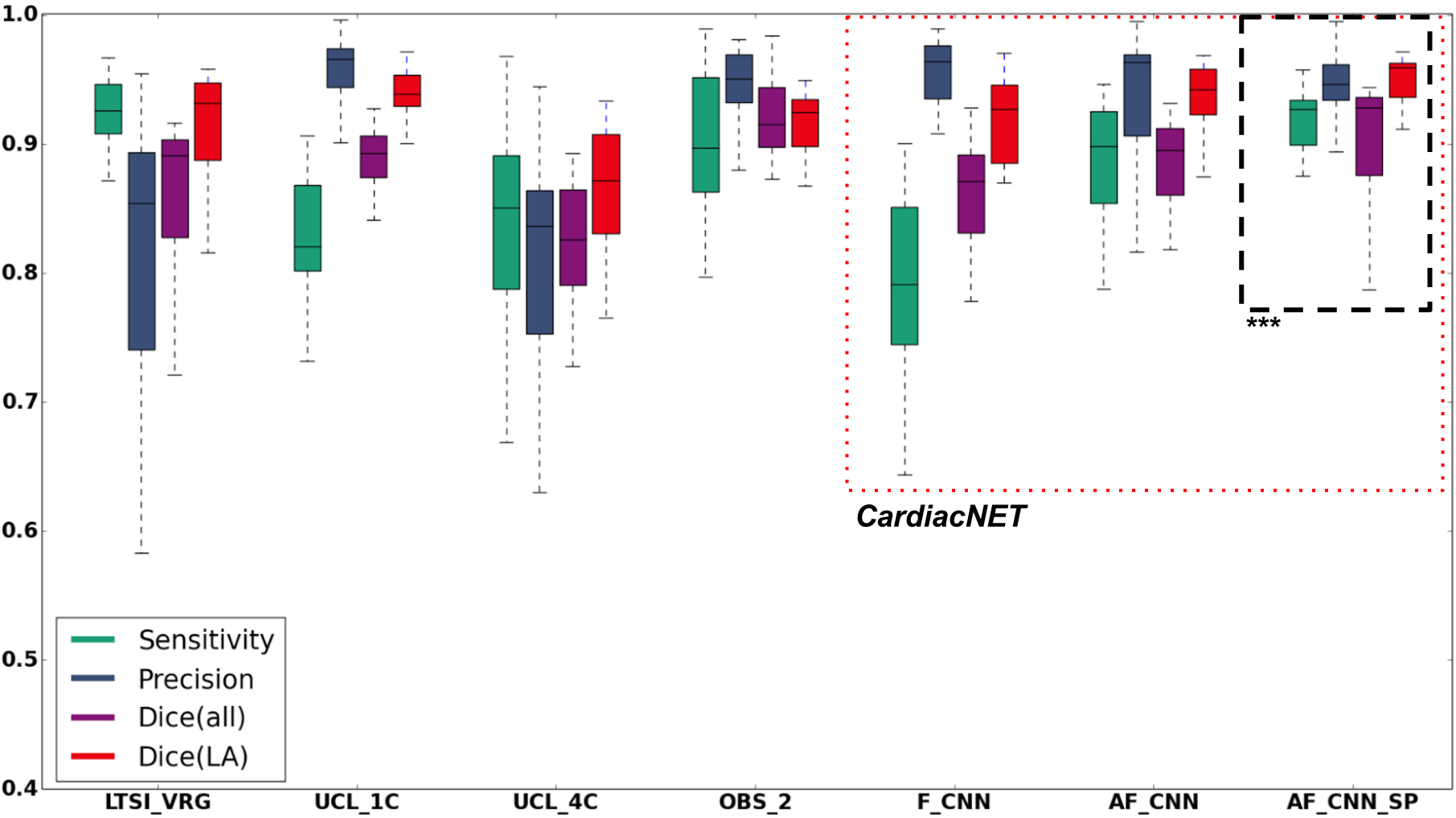}
\end{turn}
\caption{Box plots for sensitivity, precision, and Dice index for state-of-the-art (LTSI\_VRG,UCL\_1C, UCL\_4C, OBS\_2) and proposed methods (F\_CNN, AF\_CNN, AF\_CNN\_SP) on the LA segmentation benchmark}
\vspace{-0.5 cm}
\end{figure}

\section{Discussions and Concluding Remarks}
The advantage of \textit{CardiacNET} is accurate and efficient method for both LA and PPVs segmentation in atrial fibrillation patients: combined segmentation of the LA and PPVs. Precise segmentation
 of the LA and PPVs is needed for ablation therapy planning and clinical guidance in AF patients. PPVs have a greater number of anatomical variations than the LA-body, leading to challenges with accurate segmentation. Joint segmentation the LA and PPVs is even more challenging compared to sole LA-body segmentation.  
Nevertheless, with all available quantitative metrics, the proposed method has been shown to greatly improve the segmentation accuracy on the existing benchmark for LA and PPVs segmentation. The benchmark evaluation has also allowed the method and its variations to be cross-compared on the same dataset with other existing methods in literature.

%In \textit{CardiacNET} the segmentation problem of the LA is considered as a combination of both the LA-body and PPVs. It is necessary to visualize the sections of the pulmonary vein’s proximally to the LA especially in pulmonary vein isolation procedures for treating AF.
 %The large number of variations in the PPVs pose a complex and challenging segmentation problem. Therefore, jointly segmenting the LA and PPVs is more challenging compared to simply segmenting the LA body chamber, as demonstrated in the results section. Nevertheless, with all available quantitative metrics, the proposed method has been shown to greatly improve the segmentation accuracy on the existing benchmark for LA and PPVs segmentation. The benchmark evaluation has also allowed the method and its variations to be cross-compared on the same dataset with other existing methods in literature. 

Despite the efficacy of the proposed method, there are several possibilities that our work can be extended in future studies. Firstly, the new method will be tested, evaluated, and validated our in more diverse data sets from several independent cohorts, and at the different imaging resolution and noise levels, and even across different scanner vendors. Secondly, extending our framework into 4D (i.e motion) analysis of cardiac images can be possible by extending our parsing strategy.  
%for many clinical applications in cardiovascular research. This is primarily due to  the clinical and scientific need of analyzing the heart and its structures in motion. 
Thirdly,  we aim to explore the feasibility of training completely 3D cardiac MRI based on the availability of multiple GPUs, or developing sparse CNNs to alleviate the segmentation problem. Fourthly, with low-dose cardiac CT technology on the rise; it is desirable to have similar network structure trained on CT scans. This notable efficacy of the deep learning strategies presented in this work promises a similar performance on CT scans.  
%computed tomography (CT) is known with its ionizing radiation effects, CT is still being used both in clinics and research. Therefore, it will be desirable to have a similar network structure that can provide quantitative analysis of LA from CT scans. A remarkable efficacy of the deep learning strategies promises a similar performance in CT images, too.  
%Nevertheless, segmentation challenge datasets provide a controlled framework where comparison and evaluation of the proposed method can be assessed objectively

In conclusion, the proposed method has utilized the strength of deeply trained CNN to segment LA and PPVs from cardiac MRI. We have shown combining information from different views of MRI by using an adaptive fusion strategy and a new loss function improves segmentation accuracy and efficiency significantly. \\
\noindent\textbf{Acknowledgment:} Thanks to Nvidia for donating a GPU for deep learning experiments. The experiments have been conducted using Tensorflow. 

\vspace{-0.3 cm}
\bibliographystyle{IEEEbib}
\bibliography{strings,refs}

\end{document}